\documentclass[twocolumn,a4,10pt]{article}
\usepackage{iasted}
\usepackage{times}
\usepackage[dvips]{graphicx}
\usepackage{textcomp}
\begin{document}

\date{}

\title{A computational approach to the covert and overt deployment of spatial attention}

\author{
Jeremy Fix \\
\and
Nicolas Rougier  \\
Cortex\\
LORIA - Inria Lorraine\\
rue du jardin botanique\\
Villers-l{\`e}s-Nancy, France \\
email: firstname.lastname@loria.fr \\
\and Frederic Alexandre\\
}

\maketitle

\thispagestyle{empty}

\noindent
{\bf\normalsize ABSTRACT}\newline
{Popular computational models of visual attention tend to neglect the influence of saccadic eye movements whereas it has been shown that the primates perform on average three of them per seconds and that the neural substrate for the deployment of attention and the execution of an eye movement might considerably overlap. Here we propose a computational model in which the deployment of attention with or without a subsequent eye movement emerges from local, distributed and numerical computations.
} \vspace{2ex}

\noindent
{\bf\normalsize KEY WORDS}\newline
{cnft, spatial attention, saccadic eye movements, anticipation}

\section{Introduction}

We (humans) all share quite naturally the feeling of a stable world while our eyes are constantly performing saccades to actively explore the visual world. These saccades are partly motivated by the need to attend available stimuli in order to eventually decide if they might be of some interest in the near future. This corresponds to what is called overt visual attention. However, another form of visual attention that is less intuitive is the covert form of attention which does not imply a subsequent motor act toward the attended location. We believe that these two types of visual attention might result from a unique paradigm but the final stage of actually making the saccade. After a brief review of the literature introducing the main concepts, we describe a model that exhibits both covert and overt attention using homogeneous computations, the resulting sequential behavior resulting from the different informational pathways.

\section{Biological fundations}
\label{sec_biology}

\paragraph{Feature- and spatial-based visual attention.} Visual attention has at least two components \cite{ReynoldsChelazzi2004}. First it can be feature specific and the a priori knowledge of visual attributes one is looking for (such as color or orientation \cite{Wolfe2005}) provides a bias that can be observed both in psychological experiments and in physiological recordings. It can also be directed spatially with or without a subsequent motor act toward the attended location, which distinguishes the covert and overt deployment of attention in the case of saccadic eye movements. \\

\paragraph{Two streams for the processing of visual information.} In \cite{Milner1992}, the authors propose that visual information is processed along two pathways. From the occipital to the temporal lobe, along the ventral stream or the 'What' pathway, the size of the neurons' receptive field increases and their selectivity is more and more complex. This pathway is supposed to be mainly involved in the recognition of visual objects. From the occipital to the parietal lobe, along the dorsal stream or the 'Where' pathway, the neurons are significantly less selective to the visual attributes of the stimuli in their receptive field. The areas along this pathway are supposed to be involved in the representation of space, in multiple frames of reference, in order to guide further actions. Although first supposed to be independent, these two processing streams might share some projections \cite{Merigan1993}.\\

\paragraph{Linking spatial visual attention and saccadic eye movements.} The first proposal that the deployment of visual attention and the execution of a motor act might rely on the same neuronal substrate has been proposed in the premotor theory of attention \cite{Rizzolatti1987} (see also \cite{Corbetta1998}). It gains support from several psychological and physiological studies \cite{Hoffman1995,Moore2006}. In particular, \cite{Hoffman1995} evaluates the performance of human subjects in a task requiring them to detect a target at a given location presented just before executing an eye movement toward the same or a different location. They observe that the detection accuracy is higher when the site of the target to detect and the saccadic location coincide.\\

The control of voluntary saccadic eye movements involve a large set of brain areas \cite{Buttner2005}. Among them, we might distinguish the lateral intraparietal area (LIP) in the posterior parietal cortex and the frontal eye field (FEF) in the frontal cortex. \\

LIP is considered as a major oculomotor area since its lesion usually causes spatial neglect of the hemifield contralateral to the lesioned site. In \cite{Gottlieb1998}, the authors study the activity of LIP cells in order to distinguish between visual, saccade, and attention related activity. In particular, they show that the activity of LIP cells are strongly modulated by the behavioral relevance of the stimuli that occupy or will occupy the receptive of the recorded cells after a saccadic eye movement. This activity is abolished when the target is removed but the saccade is still performed leading the authors to the conclusion that LIP might represent the visual saliency of their receptive fields.\\

FEF is considered as an oculomotor area since selective microstimulations can produce saccadic eye movements toward specific locations relative to the current position of the eyes. By injecting a current below the threshold producing an eye movement, \cite{Moore2003} shows that the effects of the microstimulation on the responses of V4 neurons are similar to the effects of the covert deployment of spatial attention toward their receptive field.\\

\paragraph{Memory-related and predictive responses.} 

In the case of covert deployment of spatial attention, it is shown that attention cannot be redeployed on a previously attended location. This phenomenon, known as inhibition of return \cite{Posner1980}, is classicaly represented in computational models as a spatial memory that stores the attended locations. This memory then permits to bias the deployment of spatial attention toward novel locations. It is interesting to note that some of the areas involved in the oculomotor control also exhibit memory-related activities \cite{Umeno2001,Gnadt1988}. If there is a tight link between the deployment of attention and the control of saccadic eye movements, the memory related activities in the oculomotor areas might also contribute to the selection of the next location to attend.\\

The two oculomotor areas mentioned above (LIP and FEF) also exhibit predictive responses known as quasi-visual responses. The quasi-visual neurons have visual responses and also discharge when a saccadic eye movement will bring a stimulus inside their receptive field. The most striking property of these neurons is that they respond even if the stimulus that is brought inside their receptive field is suppressed before the execution of the eye movement. Therefore, even if no stimulus is physically present inside their receptive field, they will discharge. In \cite{Sommer2006}, the authors go even further by demonstrating that the predictive responses in FEF depend on a signal sent from the superior colliculus (SC), relayed by a nucleus in the thalamus. The inactivation of this relay disrupt the predictive responses in FEF and impairs the performance in a double-step task (which requires to perform saccadic eye movements toward two memorized targets, in the order of their presentation). This leads the author to suggest that a corollary discharge of an impending eye movement is sent from the SC to the FEF. Since FEF has both memory-related and predictive responses and receive a corollary discharge from the SC, it seems possible that FEF is involved in a working memory circuit updated on the basis of the parameters of the executed movement.

\begin{figure*}[htb]
\begin{center}
\includegraphics[width=0.7\linewidth]{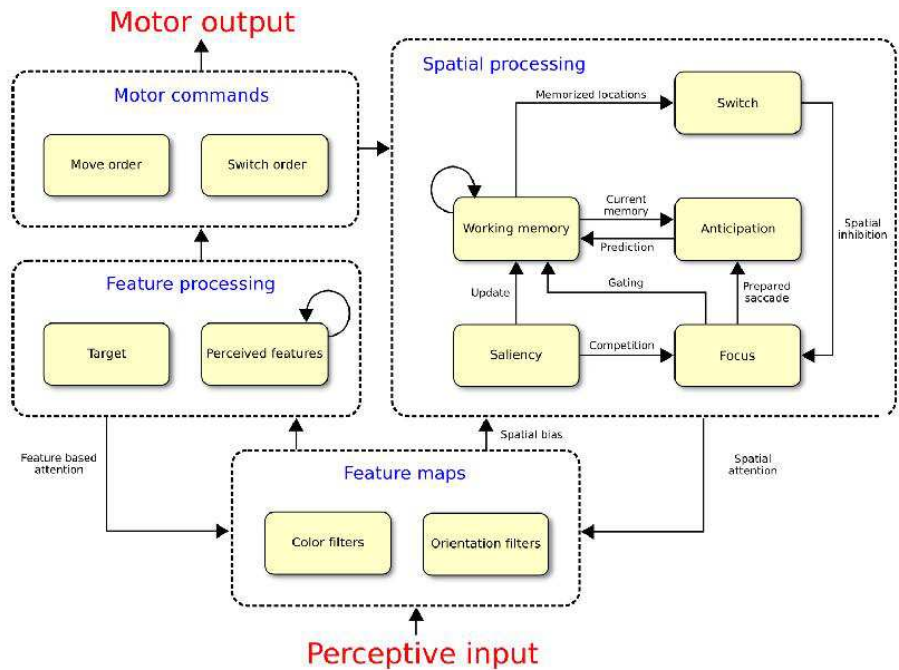}
\caption{The mechanism contains two main pathways. The first one is exclusively concerned by the processing of visual features (color, orientation) in analogy with the What pathway of the visual system of the primates. The second pathway processes the spatial information, independantly of the visual features and contains several modules involved in memorizing the spatial locations of visual targets, selecting one of them as the locus of attention or the target of a saccadic eye movement and a mechanism anticipating the consequences of an impending eye movement on the locations of the memorized targets. Please refer to the text for a more detailed description.}
\label{eps_model}
\end{center}
\end{figure*}

\section{Description of the mechanism's architecture}
\label{sec_model}

\paragraph{The framework for dynamical, distributed and numerical computations.} The proposed mechanism, built in the framework of the Continuum Neural Field Theory (CNFT) \cite{Amari1977}, considers sets of elementary units, called maps. These units interact both with lateral connections in the same map and projections from units in other maps. Each unit is characterised by its location $x$ and its activity $u(x,t)$ at time $t$, whose temporal evolution is described by the equation \ref{eq_cnft}.
\begin{equation}
 \tau.\frac{du}{dt}(x,t) = -u(x,t) + \int_y w(x,y).u(y,t) + I(x,t)
\label{eq_cnft}
\end{equation}

The lateral interaction function is a difference of gaussians which provides local excitation and distal inhibition. The input of each unit $I(x,t)$ can take two forms : it can be a simple weigthed sum of afferent units or a weighted sum of product of activities, implementing a convolution product as we will see below.\\

\paragraph{A functional description of the model.} To illustrate the way the model works, we will consider the case of a visual search task, with a limited visual field, in which the subject has to perform a saccadic eye movement toward targets defined by a set of attributes (e.g. perform a saccadic eye movement toward each red bar oriented at 45\textdegree). In this task, the basic behavior of the model is to scan sequentially the relevant locations in the visual field by deploying spatial attention until it finds a target and then to perform a saccadic eye movement toward it.\\

The global schema of the mechanism is depicted on the figure \ref{eps_model}. Each box, or module, consists of a two dimensional set of units. The visual input feeds the feature maps, where basic features (e.g. two colors and two orientations) are extracted and represented in the same eye-centered frame of reference than the input. The units in the feature maps then project along two pathways; one is feature-specific and doesn't represent the spatial location of the stimuli whereas the other only deals with spatial information independently of any feature. The clear separation between the two pathways as well as the proposition that the feature maps might represent an intermediate layer on which converge both the bottom-up signal and the top-down biases (feature- and spatial- based) are inspired from a previously proposed computational model \cite{Hamker2004}\\

Since there is both feedforward and feedback projections, a sequential description of the model is not easy\footnote{The interested reader can find videos illustrating the model on \mbox{http://www.loria.fr/$\sim$fix/}}. For the sake of simplicity, we will first describe the feature processing pathway, considering that a spatial location has been selected in the spatial processing pathway, that biases the representation in the feature maps toward the features of the attended location.\\

The feature processing pathway provides the motor units with both the current extracted features (provided by the biased feature maps) and the desired ones so that they signal whether or not the attended location contains the target. It can also send a feature bias to the feature processing pathway to increase the saliency of the locations that contain at least one target's attribute.\\

The feature maps also project along the spatial processing pathway onto the saliency map. This map represents the saliency of all the locations in the visual field. As we have seen before, since the feature processing pathway can modulate the feature maps, the activity in the saliency map will also be modulated. The attended location emerges in the focus map because of the lateral interactions that lead to a behavior similar to a winner-take-all, but in a distributed and competitive way. Whereas the units in the saliency map represent the saliency of all the potential targets of the visual field, the focus map contains only one of them. In that case, we consider that the selected location is the attended location and the focus map projects onto the feature maps in order to bias them as mentioned above.\\

Since attention is deployed sequentially until the system finds the target, the attended locations are memorized in a recurent circuit (working memory). This memory biases the competition in the focus map through the switch circuit that is also under the command of the switch order. For exemple, when the attended location does not contain the target, the switch order activates the switch mechanism which inhibits the previously attended locations. As a consequence, these locations receive a negative bias, which favours the non-previously attended locations to be selected.\\

\begin{figure*}[htb]
\begin{center}
\includegraphics[width=0.7\linewidth]{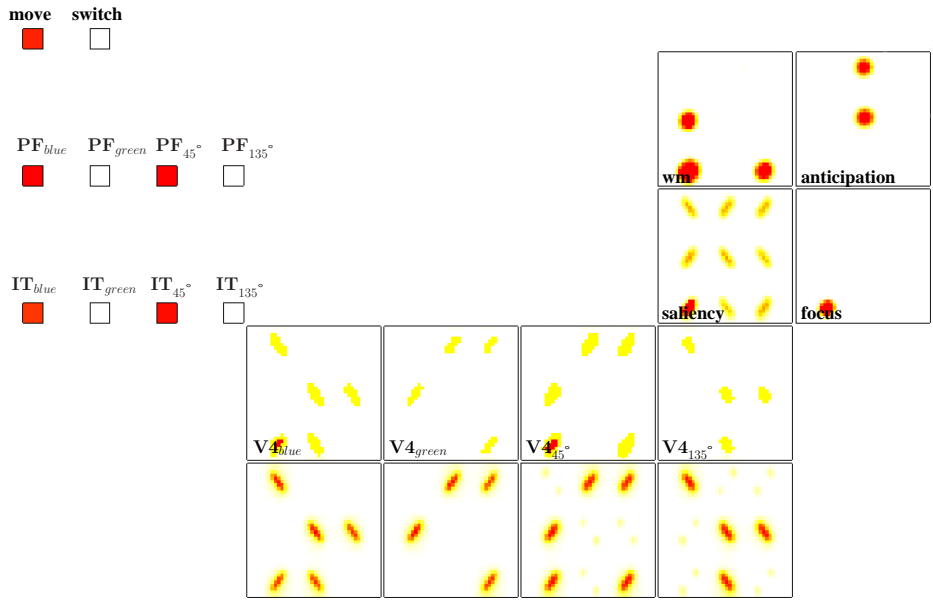}
\caption{Snapshot of the activites of the network when spatial attention has already been deployed on three spatial locations. The maps illustrated here are the same than on the figure \ref{eps_model}. The perceptive input is encoded by the maps at the bottom. They represent the result of filtering the visual input, represented on the figure \ref{eps_covert}, along four dimensions : blue, green, 45 degrees and 135 degrees. The Feature maps are made of the four V4 maps. The feature processing pathway consists of the PF and IT maps. The motor commands are represented by the {\sc move} and {\sc switch} units. Finally, the spatial processing pathway is made of the {\sc saliency}, {\sc focus}, {\sc wm} and {\sc anticipation} maps. The description of the activities in the network is given in the text.}
\label{eps_simulation}
\end{center}
\end{figure*}

The last map, that we did not describe yet, is the anticipation map. Until now we only mentionned the covert deployment of attention but as we mentionned at the beginning of the section, the system has to perform a saccadic eye movement toward the target when it finds it. The order to perform the movement is provided by the move order unit. The units in the anticipation map, which receive both the current memorized locations and the saccadic target, compute a prediction of which locations will be occupied by the memorized locations after the eye movement. The mechanism used to compute this prediction, described in more details in \cite{Fix2007Abials}, basically relies on the convolution product of the two inputs. This prediction is then combined with the input of the post-saccadic saliency map to update the memory. One reason of the updating of the memory is to avoid the system to go back and forth on two targets. It has a role similar to the inhibition of return which prevents spatial attention to be redeployed onto a previously attended location.\\

\section{Simulation and results}

\begin{figure}[htb]
\begin{center}
\fbox{\includegraphics[width=0.7\linewidth]{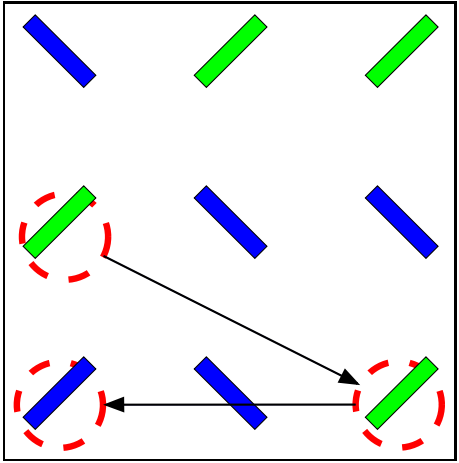}}
\caption{Scanpath of the covert deployment of spatial attention. The circles represent which locations have been attended and the arrows represent the order of the scan. We remind that the task is to find the blue target oriented at 45 degrees and perform an eye movement toward it.}
\label{eps_covert}
\end{center}
\end{figure}

The figure \ref{eps_simulation} illustrates a snapshot of the activites of the network made of several maps of $40\times40$ units. The input (blue or green bars oriented at 45 or 135 degrees) that feeds the system is shown on the figure \ref{eps_covert}. We remind that the task is to find the blue target at 45 degrees and then to perform an eye movement toward it. In a more complex scenario than the one considered here, other targets would be present in the visual field and the system should go one with this task. The activities illustrated on the figure \ref{eps_simulation} are the activites when the system has already selected spatially three targets that are stored in the working memory \textbf{wm}. Here, when we are speaking about the selected targets, we should say more accurately the regions in the visual field on which spatial attention has been deployed. These three regions are circled on the figure \ref{eps_covert}, the arrows between the circled targets representing the pathway of spatial attention.\\

We will now describe the shape of the activities in all the maps in order to clarify why these activites are like they are. The four maps at the bottom of the figure represent the result of filtering the input along the four dimensions blue, green, 45 degrees and 135 degrees. These maps feed the feature maps made of the four V4 maps. The amplitude of the activites in these maps are almost the same but two important differences should be noticed. First, the activities in V4$_{blue}$ and V4$_{45\,^{\circ}}$ are slightly more important than the activities in V4$_{green}$ and V4$_{135\,^{\circ}}$. This slight difference is a consequence of the feature based bias provided by the What pathway. This bias originates from the PF maps which encode which attributes are relevant for the task. The second difference to notice is the strong bias in amplitude for the bottom left target in the V4 maps. Here, the origin of this bias in amplitude is different from the previous case and is a consequence of a spatial based bias coming from the Where pathway. Indeed, the focus map has only one excited region that is interpreted in the model as the region in the visual field on which spatial attention is deployed. This signal projects back to the V4 maps then leading to the observed difference in amplitude in these maps.\\

The What pathway, made of the {\sc target} and {\sc perceived feature} maps on the figure \ref{eps_model}, here consist of the \textbf{PF} and \textbf{IT} maps. We already mentioned before that the \textbf{PF} maps encode the visual attribute relevant to the task. The \textbf{IT} maps integrate their inputs for the \textbf{V4} maps and represent the most saliant attributes of these maps. Since the \textbf{V4} maps are biaised in order to represent the attended location with a higher amplitude, the \textbf{IT} maps are indirectly biased to represent the visual attributes of the attended location. The two units \textbf{move} and \textbf{switch} just integrate their inputs from the \textbf{PF} and \textbf{IT} maps, which respectively represent the relevant visual attributes and the visual attributs under the locus of attention, in order to signal whether or not the stimulus under the locus of attention is the target. If it is not the target, the \textbf{switch} unit's activity is higher, whereas if it is the target, the \textbf{move} unit's activity is higher. On the illustration, it is the \textbf{move} unit that has a higher activity since the stimulus under the locus of attention is actually the target the system is looking for.\\

The Where pathway is made of four maps : \textbf{saliency}, \textbf{focus}, \textbf{wm} and \textbf{anticipation}. The \textbf{saliency} map integrates its inputs coming from the \textbf{V4} maps and then represents the saliency of all the locations in the visual field, since the \textbf{V4} maps represent an intermediate layer receiving both a feature-based and a spatial-based bias. The \textbf{saliency} map feeds the \textbf{focus} in which lateral connections ensure that a competition will select only one of the targets. The interpretation of the activities of the \textbf{focus} map is that it actually encodes the region on which spatial attention is deployed. The state of the network illustrated here is its state after several deployment of attention. The previously attended locations are encoded in the \textbf{wm} map. The three excited regions represent the two previously attended locations and the current attended one. The last map to comment is the \textbf{anticipation} map. The activities in this map should be interpreted as the future locations that the currently memorized locations should occupy after the eye movement toward the attended location, restricted to the locations that will still be in the visual field after this movement. Of the three memorized locations, only two will still be in the visual field; these two are the ones encoded in the \textbf{anticipation} map. If we would observe the dynamical behavior of the system, the next step would be for it to perform an eye movement toward the target under the locus of attention, and then go on with the task of localizing the next target. After this eye movement, the post-saccadic saliency map and the anticipation signal would be combined to update the spatial memory. Then the system would not have to scan the previously scanned locations since it would already ``knows'' that these locations do not contain a target.

\section{Discussion}

To close the loop between biological inspiration and the proposed computational model, we can provide some suggestions on the link between the maps in the model and the brain areas. The feature maps, which are the recipient of both bottom-up and top-down influences, are similar to what is known about the visual area V4. Even if it is very schematic in the model, the feature specific pathway could be compared to the neurons along the occipito-temporal pathway whose receptive field are growing in size and the selectivity is becoming more and more complex (in the model, the cells in the feature processing maps have the same selectivity, in the feature space, than the feature maps). \\

In the spatial processing part of the model, the correlation between the maps and the brain areas can be done on functional criteria rather than on anatomical ones. More specifically, the saliency map, even if the existence of only one saliency map is controversial, could be compared to LIP, as described in the section \ref{sec_biology}, with cells encoding in a eye-centered frame of reference the relevance of the locations of the visual field. The selection of a target for a saccadic eye movement might be more spread than localized in a single area. Several candidates can be considered such as the frontal eye field, the superior colliculus and also the basal ganglia, since they project onto the reticular formation (the latest set of nuclei before the ocular motoneurons) or interact with the superior colliculus which projects on it. Under the asumption that attention is deployed sequentially, there also might be a system to select the next attended location, probably based on a memory of the previously attended ones. Again, this system might involve several brain areas (e.g. LIP, FEF have both memory-related activities and dlPFC is shown to be involved in spatial short-term memory \cite{Goldman1987}) and we decided, in the model, to consider a localised circuit to model it. Finally, we also propose that the memory is encoded in a eye-centered frame of reference and that it guides the scanpath of both covert and overt attention.\\

Among the functional components of the model, three are of particular importance, namely the selection of the next attended location, the memorization of the attended locations and the anticipation of the consequences on the working memory of an impending eye movement. We believe that the neural substrate mediating these three functions could involve LIP, FEF, dlPFC, SC and the basal ganglia. During the last two decades, the basal ganglia has been shown to be involved both in cortical and subcortical loops \cite{Alexander1986,Hikosaka2000}. FEF ans SC both have saccade related activites and and interact with the basal ganglia (for exemple the SNr nucleus of the basal ganglia strongly inhibits SC) in such loops. The basal ganglia also receive afferences from many part of the cortex and is influenced by the expected or obtained reward through its dopaminergic circuits. These subcortical nuclei are then at a good position to mediate the selection of the next saccadic target. Since several recent studies are showing that the deployment of attention and the ocolomotor control might share the same substrate, the basal ganglia could also play an important role in the deployment of spatial visual attention. For the spatial memory and the capacity to anticipate the consequences of a saccadic eye movement, the works of \cite{Sommer2006} provide us with important results. They show that SC is sending a corollary discharge signal to FEF, through the mediodorsal nucleus of thalamus (MD). This nucleus is also shown to be one component of the cortico-basal loops involving the dlPFC, FEF and the basal ganglia. A temporary inactivation of MD leads to a reduction of anticipatory responses in FEF and also in impairements in double step tasks which require to memorize the targets toward which saccadic eye movements should be executed. These recent results point out the possibility that loops involving dlPFC, FEF, SC, the basal ganglia and the thalamus could play a significant role in keeping and updating a spatial working memory.\\

Whether or not to execute an eye movement is of course strongly correlated to the task the model is supposed to achieve. More specifically, the triggering of a saccade may be performed voluntarily (as in visual search for example) but may also be completely driven by the stimulus itself (e.g. reflexes) and
bypass most of the circuitry described in the first section. Hence, we do not claim to have modeled the whole attentional system but we think the proposed framework, which is strongly constrained, might help us going further in considering visual attention as an emergent property of the interaction of elementary units.

\bibliographystyle{unsrt}
\begin{footnotesize}
\bibliography{biblio}
\end{footnotesize}

\end{document}